\RequirePackage{fix-cm}
\documentclass[sigconf]{aamas} 
\usepackage{graphicx}
\usepackage{caption}
\usepackage{subcaption}
\usepackage{array}
\usepackage{amsmath}
\usepackage{amsfonts}
\usepackage{bm}
\usepackage{xcolor}
\usepackage{multirow,array}
\usepackage{comment}
\usepackage{url} 
\usepackage{tikz}
\usepackage[normalem]{ulem}
\usetikzlibrary{matrix,positioning}
\usepackage{makecell}
\usepackage[flushleft]{threeparttable}
\usepackage{multicol}

\setcopyright{ifaamas}
\acmConference[AAMAS '24]{Pre-print of paper accepted at 23rd International Conference
on Autonomous Agents and Multiagent Systems (AAMAS 2024)}{May 6 -- 10, 2024}
{Auckland, New Zealand}{N.~Alechina, V.~Dignum, M.~Dastani, J.S.~Sichman (eds.)}
\copyrightyear{2024}
\acmYear{2024}
\acmDOI{}
\acmPrice{}
\acmISBN{}

\acmSubmissionID{1169}

\title[Utility-based RL]{Utility-Based Reinforcement Learning: Unifying Single-objective and Multi-objective Reinforcement Learning}

\author{Peter Vamplew, Cameron Foale}
\affiliation{
  \institution{Federation University Australia}
  \city{Ballarat}
  \country{Australia}}
\email{p.vamplew,c.foale@federation.edu.au}

\author{Conor F. Hayes}
\affiliation{
  \institution{Lawrence Livermore National Laboratory}
  \city{Livermore}
  \country{USA}}
\email{hayes56@llnl.gov}

\author{Patrick Mannion, Enda Howley}
\affiliation{
  \institution{University of Galway}
  \city{Galway}
  \country{Ireland}}
\email{patrick.mannion@universityofgalway.ie}
\email{enda.howley@universityofgalway.ie}

\author{Richard Dazeley, Scott Johnson}
\affiliation{
  \institution{Deakin University}
  \city{Geelong}
  \country{Australia}}
\email{r.dazeley,s223861809@deakin.edu.au}

\author{Johan Källström}
\affiliation{
  \institution{Linköping University}
  \city{Linköping}
  \country{Sweden}}
\email{johan.kallstrom@liu.se}

\author{Gabriel Ramos}
\affiliation{
  \institution{Universidade do Vale do Rio dos Sinos}
  \city{S\~ao Leopoldo}
  \country{Brazil}}
\email{gdoramos@unisinos.br}

\author{Roxana R\u{a}dulescu, Willem R\"opke, Diederik M.\ Roijers}
\affiliation{
  \institution{Vrije Universiteit Brussel}
  \city{Brussels}
  \country{Belgium}}
\email{roxana.radulescu,willem.ropke,diederik.roijers@vub.be}

\begin{abstract}
Research in multi-objective reinforcement learning (MORL) has introduced the \emph{utility-based} paradigm, which makes use of both environmental rewards and a function that defines the utility derived by the user from those rewards. In this paper we extend this paradigm to the context of single-objective reinforcement learning (RL), and outline multiple potential benefits including the ability to perform multi-policy learning across tasks relating to uncertain objectives, risk-aware RL, discounting, and safe RL. We also examine the algorithmic implications of adopting a utility-based approach.
\end{abstract}


\newcommand{\BibTeX}{\rm B\kern-.05em{\sc i\kern-.025em b}\kern-.08em\TeX}

\DeclareSymbolFont{bbold}{U}{bbold}{m}{n}
\DeclareSymbolFontAlphabet{\mathbbold}{bbold}

\definecolor{darkyellow}{RGB}{64,64,0}

\definecolor{luisas_color}{HTML}{5c4ae4}
\definecolor{enda_color}{HTML}{E44A60}

\definecolor{fredrik_color}{HTML}{234567}

\begin{document}


\pagestyle{fancy}
\fancyhead{}


\maketitle 


\section{Introduction}\label{sec:intro}
Multi-objective reinforcement learning (MORL) has emerged as an important sub-field of reinforcement learning (RL) research \cite{roijers2013survey,hayes2022practical}. So far the flow of knowledge has primarily been from conventional single-objective RL (SORL) into MORL, with algorithmic innovations from SORL being adapted to the context of multiple objectives \cite{yang2019generalized, chen2019meta,alegre2022optimistic,reymond2023actor}. This paper runs counter to that trend, as we will argue that the utility-based paradigm which has been widely adopted in MORL \cite{cai2023distributional,hayes2022practical,ruadulescu2020multi}, has both relevance and benefits to SORL. We present a general framework for utility-based RL (UBRL), which unifies the SORL and MORL frameworks, and discuss benefits and potential applications of this for single-objective problems -- in particular focusing on the novel potential UBRL offers for applying \textit{multi-policy learning} to MDPs, thereby allowing greater flexibility and decision-maker control over the behaviour of agents. We also highlight lessons learned from MORL research regarding the algorithmic implications of a utility-based approach as a guide to future single-objective UBRL research and practice.

\section{Formalising utility-based RL}\label{sec:formalities}
\subsection{MDPs, MOMDPs and Optimisation criteria}\label{sec:mdps&criteria}

SORL and MORL share a common foundation in the assumption that the environment can be represented as some form of Markov Decision Process (MDP). In the single objective case, a MDP is represented by the tuple $\langle S, A, T, \gamma, \mu, R \rangle$, where: 
\begin{itemize}
    \item $S$ is the state space
    \item $A$ is the action space
    \item $T \colon S \times A \times S \to \left[ 0, 1 \right]$ is a probabilistic transition function
    \item $\gamma \in [0, 1)$ is a discount factor
    \item $\mu \colon S \to [0,1]$ is a probability distribution over initial states 
    \item $R \colon S \times A \times S \to \mathbb{R}$ is a scalar-valued reward function
\end{itemize}

For a single-objective problem of this nature, the agent's aim is to discover a policy $\pi$ which maximises either the finite-horizon undiscounted reward (Equation \ref{eq:finite-sum}), the infinite-horizon cumulative discounted reward (Equation \ref{eq:infinite-sum}), or the average reward (Equation \ref{eq:ave-reward}). We note that the choice of criteria to optimise is a decision made by the system designer, rather than being an inherent property of the MDP itself.

\begin{equation}
    V_\pi = \mathbb{E} \left[ \sum\limits^{N-1}_{i=0} r_i \:\middle|\: \pi, \mu \right]
    \label{eq:finite-sum}
\end{equation}

\begin{equation}
    V_\pi = \mathbb{E} \left[ \sum\limits^\infty_{i=0} \gamma^i r_i \:\middle|\: \pi, \mu \right]
    \label{eq:infinite-sum}
\end{equation}

\begin{equation}
    V_\pi = \lim_{N\rightarrow\infty} \frac{1}{N} \mathbb{E} \left[ \sum\limits^{N-1}_{i=0} r_i \:\middle|\: \pi, \mu \right]
    \label{eq:ave-reward}
\end{equation}

MORL differs from SORL in being based on the concept of a Multi-Objective Markov Decision Process (MOMDP). This is identical to a MDP in all respects other than the reward, where rather than a scalar reward $R \colon S \times A \times S \to \mathbb{R}$, the MOMDP has a vector-valued reward function $\mathbf{R}: S \times A \times S \to \mathbb{R}^d$ which specifies the immediate reward for each of the $d$ objectives.

The optimisation criteria in Equations \ref{eq:finite-sum}--\ref{eq:ave-reward} are insufficient for a MOMDP, as they only provide a partial-ordering over the vector values. Therefore the utility-based paradigm for MORL assumes the existence of a \emph{utility function} (sometimes called a \emph{scalarisation function}) $u$, which maps the multi-objective value of a policy to a scalar value \cite{hayes2022practical}. Note that the details of this utility function may or may not be available in advance -- in the latter case the agent may need to identify a set of policies which would be optimal under different parameterisations of the utility function. We will denote a parameterised utility function as $u_\omega$.

Given the existence of $u$ we can define two alternative optimisation criteria, depending on the stage at which $u$ is applied (the reasoning for this distinction will be discussed further in Section \ref{sec:implications})\footnote{Note that these equations relate to the infinite-horizon discounted return which has been the most widely studied criteria in the MORL literature. Similar utility-based equations can readily be defined for the finite-horizon and average-reward criteria.}. Equation \ref{eq:ser} defines the scalarised expected reward (SER) where $u$ is applied to the expected vector reward, whereas Equation \ref{eq:esr} defines the expected scalarised reward, where $u$ is applied inside the expectation operator. In either case, these equations provide a scalar value which defines a total ordering over policies.

\begin{equation}
    V_{u}^{\pi} = u\left(\mathbb{E} \left[ \sum\limits^\infty_{i=0} \gamma^i {\bf r}_i \:\middle|\: \pi, \mu \right]\right)
    \label{eq:ser}
\end{equation}

\begin{equation}
    V_{u}^{\pi} = \mathbb{E} \left[ u\left( \sum\limits^\infty_{i=0} \gamma^i {\bf r}_i \right) \:\middle|\: \pi, \mu \right]
    \label{eq:esr}
\end{equation}

\subsection{Utility-based RL as a general framework}

In the previous subsection we highlighted the differences between conventional single-objective RL and utility-based MORL. We now consider the commonalities between these approaches, and argue that utility-based RL provides a general framework which encompasses both SORL and MORL\footnote{The name `\emph{utility-based} is not intended to imply that standard SORL is \emph{not} maximising a measure of utility, but to emphasise our framework's \emph{explicit} consideration of user-utility. In standard SORL the choice of optimisation criteria is made by the developer whereas in multi-policy UBRL the final decision rests with the end-user.}.

We start with the straightforward observation that MDPs are a subclass of MOMDPs. Any MDP can be mapped to an equivalent MOMDP with a reward function Which consists of a one-dimensional vector $\mathbf{R}=[R]$.

Having done this conversion, then utility-based methods such as those defined in terms of the optimisation criteria in Equations \ref{eq:ser} and \ref{eq:esr} can be applied to the resultant MOMDP. More generally we might consider any utility function which maps from the expected mean rewards or the expected distribution of rewards to a scalar value -- we will explore examples of this in Sections \ref{sec:risk-aware} and \ref{sec:discounting}. In addition the utility function may itself be parameterised, so that a set of policies would be required in order to optimise the criteria across all possible instantiations of $u$.



We note that some authors have previously criticised approaches such as MORL and risk-aware RL as ``special cases'' \cite{silver2021reward}, and we anticipate that similar criticisms may be levelled at UBRL. To the contrary, we argue that the UBRL framework presented here is a strict generalisation of prior RL frameworks. It can encompass standard scalar-reward RL by simply setting the number of objectives $n$=1 and the utility function $u$ to the identity function, while also supporting other more complex forms of RL, such as MORL.

\section{Motivation for utility-based SORL}\label{sec:utility-based-sorl}

While it is clear that SORL problems can be represented within the UBRL framework, the question remains as to whether this is an unnecessary complication, or whether it provides tangible benefits.

One advantage of a utility-based paradigm is that it can simplify the task of reward engineering \cite{dewey2014reinforcement} -- that is, designing rewards in such a way that maximising them induces the desired behaviour from an RL agent. In many cases it may be easy to identify events (significant changes in state) within the MDP for which rewards should be provided (e.g. picking up an object). But the task of tailoring the magnitude of those rewards and the choice of optimisation criteria to engender the desired behaviour is non-trivial (see for example \cite{knox2023reward}). Consider an environment with gold nuggets of differing size at varying distances from the starting location. Whether an agent prioritises collecting nearby low-value objects over more distant, higher-valued objects will depend both on the relative magnitude of the rewards and factors of the optimisation criteria, such as the choice of discounting parameter. The utility-based framework separates the specification of the rewards within the MDP (i.e. defining the environment) from the definition of utility (i.e. defining the desired outcome), which may make the reward engineer's task easier.

This is particularly true when considered in light of experience in MORL research which has found that a key benefit of a utility-based approach is that it allows for the creation of multi-policy learning algorithms \cite{hayes2022practical}. Rather than aiming to learn a single policy optimal for a specific definition of utility, multi-policy algorithms instead consider a set of possible utility functions $U=\{U_1, .., U_m\}$, and produce a \emph{coverage set} of policies such that the agent has an optimal policy for every member of $U$.

Learning multiple policies in this way is not possible within conventional SORL, and so represents a novel contribution of the UBRL framework that greatly enhances the RL process. We can define an MDP with a simple reward structure, learn multiple policies for that MDP based on a variety of different definitions of utility, and then select the utility function which produces the most desirable outcomes. This will be significantly easier than trying to specify \emph{a priori} the rewards required to produce the desired behaviour. It also means that the final decision can be left in the hands of the ultimate user of the system, rather than inappropriately requiring it to be made by the system engineers \cite{vamplew2022scalar}. In addition this provides flexibility should the desired behaviour change over time, as a new policy can be selected without any need for further learning.

The benefits of multi-policy learning may also be realised at relatively little additional cost. The utility-based approach allows for \emph{inner-loop} multi-policy methods, in which the multiple policies are learned in parallel \cite{hayes2022practical}. For example, the Conditioned Network algorithm trains a single Deep Q-Network, conditioned on both the current MOMDP state and the values of the parameters $\omega$ of the utility function. Experiences gained while following the policy for one value of $\omega$ are leveraged to update Q-values for policies conditioned on different values of $\omega$. This greatly increases the sample-efficiency of these methods. 
\section{Potential single-objective applications of utility-based RL}\label{sec:applications}

In this section we will discuss several potential applications of the utility-based paradigm within SORL. These are intended as representative examples where UBRL provides benefits, rather than an exhaustive list. These represent both a reframing of existing concepts such as risk-aware RL to the UBRL framework, as well as more novel concepts made possible by the adoption of this framework.

\subsection{Multi-policy methods for hard-to-define objectives}
Consider the scenario of a mining company which has the choice between carrying out its usual operations or following a riskier course of action that can potentially lead to a lot of the resource being mined (e.g., committing most of the workforce to exploratory excavation). The company has outstanding contracts that commit it to sell a given amount to certain costumers at a set price and leading to an incurred penalty if these amounts are not delivered.

There are various ways such a scenario might be addressed using standard RL approaches:
\begin{itemize}
    \item We might model this as a constrained RL task \cite{gattami2021reinforcement}, where the fulfillment of contracts is treated as an inviolable constraint. But perhaps this is overly restrictive, as the company management may be prepared to breach contracts if the potential pay-off is sufficiently high? 
    \item It could be modelled as a SORL problem, with the value of the resource mined represented as a positive scalar reward, and the penalty for unfulfilled contracts as a negative reward. This would allow the agent to select the riskier action if the potential gain offsets the contractual penalty. But such an approach might fail to account for longer-term reputational damage arising from unfulfilled contracts.
    \item MORL methods could model the reputational harm as an additional objective to be minimised. But this may be very difficult to define quantitatively.
\end{itemize}

As a result, the preferred approach of the company's management might be to generate a set of alternative policies using UBRL, and make an \emph{a posteriori} decision about which policy to follow. This could be accomplished by defining the reward in terms of the quantity of the resource mined, and then specifying the utility function as in Equation \ref{eq:mining-utility}, where  $d$ is the monetary value of each unit of the resource, $c(\cdot)$ is a binary function returning 1 if the contract terms are breached and 0 otherwise, $p$ is the penalty for breaching the contract, and $h$ is an estimate of the financial impact of the reputational harm caused by a contract breach.

\begin{equation}
    u_h = d * \sum\limits^{N-1}_{i=0} r_i - c(\sum\limits^{N-1}_{i=0} r_i) * (p + h)\\
    \label{eq:mining-utility}
\end{equation}

Note that the utility term $u$ is conditioned on the value of $h$. Rather than learning an optimal policy for a single estimate of cost of the reputational harm, this utility-based formulation enables the UBRL agent to learn multiple policies, each optimal with respect to a different estimated value of $h$. The multiple policies produced via this process would then be presented to the management, allowing them to make an informed decision about the best policy to execute. Existing multi-policy MORL methods such as the Conditioned Network approach of \cite{abels2019dynamic} could readily be adapted to this task by conditioning Q-values and policies on $h$.

\subsection{Multi-policy risk-aware RL}\label{sec:risk-aware}
Many authors have investigated risk-aware RL agents -- that is, agents which apply some awareness of risk during their decision-making rather than simply aiming to maximise the expected return. A common approach to address this problem is to use a distributional form of RL \cite{bellemare2023distributional,hayes2021expected,nguyen2021distributional, martin2020stochastically}, which learns the distribution of future returns from any state rather than just the mean expected return. These approaches can take into account various aspects of this distribution when identifying the optimal policy (e.g. best worst case, more than 60\% chance of profit, etc.) \cite{cheng2023distributional,greenberg2022efficient,yang2023safety, fawzi2022discovering, hayes2023monte}.

Again we would argue that such an approach is essentially already utility-based. In this case the choice of constraints applied to the distribution represents the definition of the utility function. Conventional SORL approaches to risk-awareness assume that these constraints are fixed \emph{a priori}, and learn a single policy which is optimal under those constraints. 

In contrast by adopting the UBRL mindset, we can envisage an algorithm which learns, in parallel using multi-policy methods, a set of policies which are optimal under different risk-sensitivity preferences. These could then be presented to a human decision-maker for selection of the policy to actually be executed.

For example, consider the conditional
value at risk (CVaR) defined in Equation \ref{eq:cvar}, which has been widely-used as an optimisation criteria in risk-aware RL \cite{singh2020improving}. Here $Z$ represents the distribution of future returns, and $\alpha\in[0,1]$ is a parameter controlling the risk-sensitivity of the agent. In conventional risk-aware RL, the value of $\alpha$ is fixed and a single optimal policy is found. If we instead treat CVaR$_\alpha$ as a utility function parameterised by $\alpha$, we can apply multi-policy methods to find policies with a diverse range of different sensitivities to risk, and then allow a human decision-maker to select their preferred policy.

\begin{equation}
    \text{CVaR}_\alpha(Z) = \mathbb{E} \left[ Z \:\middle|\: Z\leq\text{VaR}_\alpha(Z) \right]
\label{eq:cvar}
\end{equation} 

\subsection{Multi-policy discounting}\label{sec:discounting}

We noted earlier in Section \ref{sec:formalities} that the choice of optimisation criteria used in SORL (undiscounted sum, discounted sum, or average reward) is made by the system designer/user rather than being an inherent property of the MDP itself. As such, this decision could in itself be regarded as a definition of user utility, although it usually is not framed as such. Similarly we would contend that the discounting term $\gamma$ used in Equation \ref{eq:infinite-sum} can be viewed as a form of utility definition. This term does not originate from the MDP itself, and two agents using different discounting rates may well derive differing optimal policies from the same MDP. Conventional SORL requires that the discounting rate be fixed prior to learning, and the impact of this choice on the final policy is not evident to the user. The only way to gain insight into the effect of the discounting rate on the agent's behaviour is to subsequently train another agent on the same MDP using a different value for $\gamma$ which is inefficient.

A superior approach is to treat the discounting rate as a parameter of the utility function as shown in Equation \ref{eq:u-gamma}, and use a multi-policy UBRL algorithm to simultaneously learn optimal policies for a range of different values for $\gamma$. The resulting policies can then be presented to the user, allowing them to make a fully informed choice of the appropriate policy\footnote{We note that the utility definition in Equation \ref{eq:u-gamma} differs from both forms of utility previously used in MORL (Equations \ref{eq:ser} and \ref{eq:esr}), in that it is the per-time-step reward, rather than the summed return, that is being transformed.} 

\begin{equation}
    u_\gamma = \sum\limits^\infty_{i=0} \gamma^i r_i
    \label{eq:u-gamma}
\end{equation}

\subsection{Satisficing agents} \label{sec:satisficing}
AI safety researchers have argued that `hard optimizers' like  conventional RL can be unsafe, as their absolute focus on maximising the given reward signal may lead to adverse side-effects if the rewards are not fully aligned with the actual desired behaviour \cite{taylor2016quantilizers,vamplew2018human}. This has lead to interest in developing \emph{satisficing agents} which are not incentivised to over-optimise \cite{hajiabolhassan2023online}. For example, in a multi-agent resource-gathering task we might wish to disincentivise an agent from collecting more resources than it actually needs to avoid adversely impacting other agents. In the context of MORL satisficing has been implemented by causing the agent to switch emphasis between objectives once a suitable level of return has been achieved for each objective. This can either use a hard threshold for each objective as in \cite{vamplew2021potential} or a non-linear utility which reduces the weighting for objectives as higher returns are achieved \cite{smith2023using}.

These approaches are not directly applicable to SORL due to the lack of alternative objectives. For example, reducing the gradient of the utility function as in \cite{smith2023using} would not disincentivise the agent from still trying to maximise it. So for safe SORL we might need to consider the concept of a non-monotonic utility function where the utility actually falls in value after a satisfactory amount of reward has been received, as shown in Equation~\ref{eq:nonmonotonic}. 

\begin{equation}
    u_\omega = -\lvert \omega - \sum\limits^\infty_{i=0} \gamma^i r_i \rvert
    \label{eq:nonmonotonic}
\end{equation}

This could of course also be achieved via changing the reward definition within the MDP. However that approach would not be amenable to multi-policy learning whereas a multi-policy UBRL algorithm could simultaneously learn optimal policies for various values of the threshold $\omega$, allowing insight into the appropriate value of $\omega$ which will produce the desired safe behaviour while still performing to a satisfactory level in terms of the actual task.  

\section{Implications of non-linear utility}\label{sec:implications}

The benefits of utility-based learning described earlier do not come without some additional considerations. Research in utility-based MORL has identified several issues which need to be taken into account when designing and applying UBRL methods, particularly when using non-linear forms of $u$, such as that in Equation \ref{eq:nonmonotonic}.\footnote{While MORL research has considered both linear and non-linear utility functions, we anticipate that single-objective UBRL will almost exclusively use non-linear forms as applying a linear function to a scalar reward merely scales the range of the reward without having any effect on the optimal policy.}

One fundamental issue relates to the selection of the optimisation criteria. For tasks where we care about the outcome on a per-episode basis then the Expected Scalarised Return (ESR) defined in Equation~\ref{eq:esr} is most appropriate, whereas for problems where we care about the average outcome over multiple episodes then Scalarised Expected Return (SER, Equation~\ref{eq:ser}) is the correct criteria\footnote{The distinction between ESR and SER does not arise in SORL or in UBRL with linear $U$ as the values of Equations \ref{eq:ser} and \ref{eq:esr} are equal under those settings.}. Several researchers have reported that different algorithms may be required to address the ESR and SER settings \cite{ruadulescu2020utility,roijers2018multi,vamplew2022impact}.

A non-linear utility function has further implications for value-based UBRL algorithms, as it means the returns are no longer additive which is not compatible with the Bellman equation~\cite{roijers2013survey}. To address this, UBRL algorithms based on temporal difference approaches may need to use an \emph{augmented state} which concatenates the environment state with a history of prior received rewards \cite{vamplew2022impact}.

One further issue to consider is the implementation of reward shaping. In SORL, shaping rewards are simply added to the MDP rewards prior to presenting the reward to the agent. As long as certain conditions are met, the optimal policy is not altered by the inclusion of these shaping rewards \cite{ng1999policy}. However this may no longer be the case once a non-linear utility function is applied to these combined reward values. To allow the agent to appropriately evaluate the true utility, it may prove necessary to treat the shaping reward as a separate objective~\cite{brys2014multi}.

\section{Conclusion}

The utility-based RL framework presented here unifies the previously disparate areas of single-objective RL and multi-objective RL. The utility-based approach increases the flexibility of SORL agents by facilitating multi-policy learning. This is the main novel capability of UBRL compared to standard SORL, and increases the control human decision-makers have over the RL agent, by allowing them to make an informed selection of their preferred agent after seeing which alternative behaviours exist, rather than making \emph{a priori} decisions regarding the reward design and optimisation criteria and hoping these achieve the desired outcome. UBRL may also have benefits in terms of simplifying the task of reward design. We have presented several motivating cases where multi-policy UBRL approaches have advantages over standard SORL -- uncertain objectives, risk-aware RL, discounted returns, and satisficing agents. Finally we believe that wide-spread adoption of UBRL as a unified framework would facilitate faster and easier transfer of novel ideas, algorithms and software implementations between the previously somewhat disconnected fields of SORL and MORL.




\bibliographystyle{ACM-Reference-Format} 
\bibliography{references.bib}

\end{document}